\documentclass[letterpaper, 10 pt, conference]{ieeeconf}  

\IEEEoverridecommandlockouts                              

\usepackage{amsmath,amsfonts,amssymb}
\usepackage{algorithmic}
\usepackage{algorithm}
\usepackage{array}
\usepackage[caption=false,font=normalsize,labelfont=sf,textfont=sf]{subfig}
\usepackage{textcomp}
\usepackage{stfloats}
\usepackage{url}
\usepackage{verbatim}
\usepackage{graphicx}
\usepackage{cite}
\usepackage{color}
\usepackage{xcolor}
\usepackage{tikz}
\usepackage{subfig}

\title{\LARGE \bf
Modeling and Simulation of an Elastic Passive Joint for Non-flipping Jumping Robot
}
\author{Qi Li, Liang Peng, Zhiyuan Wu, Pengda Ye, Weitao Zhang, Yi Xu and Qing Shi, $Senior Member, IEEE$ 
\thanks{*This work was supported by the National Natural Science Foundation of China under Grant U2013208 and 62088101. $ ( Corresponding $ $author:$ $Qing$ $Shi.)$  }
\thanks{Qi Li, Liang Peng, Zhiyuan Wu, Pengda Ye, Weitao Zhang, Yi Xu and Qing Shi are with Intelligent Robotics Institute, School of Mechatronical Engineering, Beijing Institute of Technology,
Beijing 100081, China and with the Key Laboratory of Biomimetic Robots and Systems, Beijing Institute of Technology, Ministry of Education, Beijing 100081, China.
        {\tt\small 3220230236@bit.edu.cn}}%
}

\begin{document}

\maketitle
\thispagestyle{empty}
\pagestyle{empty}

\begin{abstract}
To improve the environmental adaptability of small-sized robots, jumping is widely used to achieve high moving efficiency and obstacle surmounting ability. However, the flipping problem commonly existing in jumping robots will limit their application prospects. Instead of attaching additional control mechanisms and actuators, we propose an elastic passive joint (EPJ) to prevent the robot from flipping. By integrating a revolute joint, a switch and a spring at the root of the hindleg, the leg can rotate around the body during take-off phase, in which the large angular velocity will be inhibited due to the absorbed angular kinetic energy by the spring. Then, dynamic modeling and series of simulation are carried out to optimize the position and stiffness of EPJ. The simulation results reveal that, with EPJ worked, the zero-point of the angular velocity will appear when changing the position of the axis slightly. Meanwhile, the optimal spring stiffness of 1566 N/m will ensure a non-flipping jumping motion, in which the jumping height is reduced but the jumping distance is improved.

\end{abstract}

\section{INTRODUCTION}

For small-sized robots, jumping is a moving method with unique advantages which can help them overcome obstacles several times their own sizes \cite{b1}. Compared with crawling robots, jumping robots can better adapt to unstructured terrains and possess higher movement efficiency in various environments \cite{b2}\cite{b3}. However, flipping is a common problem for jumping robots. Limited by the robot’s size and quality, it is difficult to add more control mechanisms and actuators to them. Hence, achieving non-flipping jumping motion without adding extra mechanisms for control is still challenging.

One traditional method to prevent the robot from fipping is to control the position of the center of mass (CoM) \cite{b4}. During mechanism design, the optimal position of CoM and initial orientation can be obtained through dynamic analysis to reduce the possibility of flipping \cite{b5}\cite{b6}. Moreover, using sensors to detect the orientation of the robot then adjusting which through feedback control is also an important method to prevent the robot from flipping \cite{b7}\cite{b8}\cite{b9}. Another method is to add stabilizing devices to avoid flipping. To  achieve a stable landing, a momentum wheel based on the guinea fowl passive hind toe model can be attached \cite{b10}.Also, various types of tails \cite{b11}\cite{b12}\cite{b13} to adjust the aerial posture have widely been used. However, the addition of the tail increased the size of the robot, and the actuator within the mechanism also greatly increased the mass of the robot. 

In recent years, researchers have proposed many novel methods to prevent the robot from flipping. A deadbeat foot placement hopping controller has been developed using a third order Taylor series approximation to an offline dynamic model, which ensures precise foot placement on jumping trajectories\cite{b14}. Furthermore, adaptive control method \cite{b15}\cite{b16}\cite{b17} and machine learning algorithms \cite{b18}\cite{b19} are also well used in jumping robots, which can identify the status of the robot and adjust its posture during aerial phase, or achieve recovery after landing. However, these methods usually require additional sensors or actuators to adjust the robot's orientation, which increases the quality of the robot. Meanwhile, complex control algorithms also increase the difficulty of robots’ jumping control within an extremely short time.

In order to solve the problem of non-flipping jumping while ensuring the lightweight of the robot, this paper proposes an elastic passive joint (EPJ) only composed of a revolute joint, a switch and a spring. Then performs dynamic analysis on the take-off and aerial phases of the modified robot. In order to explore the effect of EPJ, a simulation based on Matlab is conducted on the jumping process of the robot to optimize the mechanism parameters.  The simulation results reveal that, the robot will flip forward without EPJ. With EPJ worked, the angular velocity can reach zero by changing the position of the axis. At the same time, the optimal spring stiffness is 1566N/m, which ensures a non-flipping jumping motion. 

\section{DESIGN AND MODELING OF THE MECHANISM}\label{design}

In the previous research, we adopted the Stephenson six-bar mechanism for biomimetic jumping and designed a robot based on single-degree-of-freedom (SDOF) locust-inspired jumping leg \cite{b20}. However, the previous research placed extremely strict requirements on the position of the CoM in order to prevent the robot from flipping. For further improving the robot's environmental adaptability, additional motion actuators need to be added, thereby changing the robot's CoM. In order to overcome the robot flipping caused by the change of the CoM, we propose a mechanism EPJ, so that the robot can still jump without flipping when the CoM changes.

\subsection{Mechanism Design}

In this section, based on Stephen's six-bar locust-inspired jumping leg, we added a revolute joint, an elastic element and a pair of switch at the root of the leg. The overall structure of the robot is shown in Figure 1.

\begin{figure}[htbp]
	\centering{\includegraphics[width=0.5\textwidth]{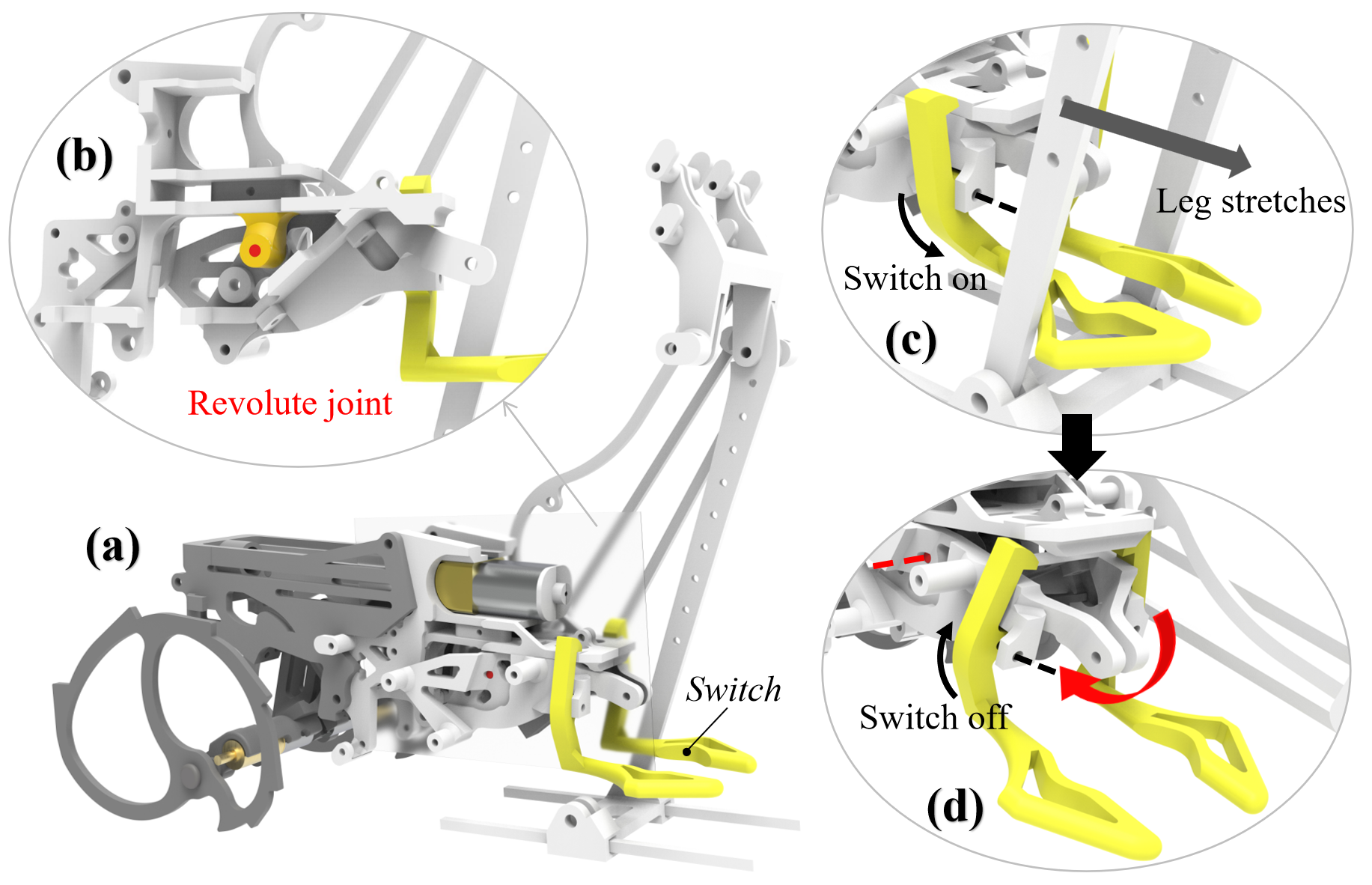}}
	\caption{Jumping robot mechanism design: (a)The rendering of the robot as a whole; (b)The longitudinal cataway view of EPJ; (c)The leg squeezes the switch to rotate it, releasing EPJ; (d)The leg gradually rotates to away from the body.}
	\label{fig}
\end{figure}

 The leg is connected to rear frame through revolute joint. Rear frame can rotate about point $O$ relative to body, and the spring is connected at both ends between the body and the rear frame. Switch is fixed on the rear frame and has two forms: switch on and switch off. The mechanism composed of a revolute joint $O$, a spring, a switch, and the rear frame enables the robot to jump smoothly, which we call EPJ.

EPJ has two operational modes: open and closed, and the switching between the two operational modes is controlled by the spring-switch mechanism. As shown in Fig.1(c) and Fig.1(d), the operational process of EPJ can be summarized as follows:

\begin{enumerate}
	\item Triggering: Before the jump, the switch is in a locked state, which can be regarded as EPJ being fixed to the body. The moment the jump begins, the leg pushes back and squeezes the switch, causing the switch to open and releasing EPJ.
	\item Operation: Before the robot leaves the ground, the ground exerts a force $F_G$ on the robot as a whole. During the take-off process, point $O$ generates a clockwise moment $M_{B}$ on the body and a counterclockwise moment $M_{A}$ on the leg. Therefore, EPJ rotates around the axis $O$, while spring $B$ is stretched.
	\item Invalidating: After the spring of EPJ is stretched to its limit, it begins to contract. Under the tension of the spring, the leg gradually rotates away from the body. During this process, the switch collides with the body, causing the switch to relock. After the switch is locked, the robot can be regarded as a rigid body with no relative motion inside and performs oblique throwing motion.
\end{enumerate}

\subsection{Modeling of The Jumping Process}

The purpose of this section is to analyze how EPJ improves the jumping orientation of the robot. To better analyze the working principle of EPJ, the following assumptions are made:

\begin{enumerate}
	\item Point $G$ and the ground are in rigid contact with no elastic deformation.
	\item There is only static friction between point $G$ and the ground, and no sliding friction. In other words, the leg rotates around the axis located at point $G$ as a fixed axis rotation.
	\item When the robot is off the ground, the leg is stationary relative to EPJ, which can be regarded as a mechanism. This mechanism rotates relative to the body around axis $O$.
\end{enumerate}

The jumping process of the robot is primarily divided into two stages: the takeoff stage and the aerial stage. In the previous research, we conducted a detailed dynamic and kinematic modeling of the jumping robot's takeoff stage using the D-H parameter method and Lagrangian equations. By calculation, we obtained the leg and body orientations at any moment before leaving the ground. Therefore, this paper focuses on analyzing the influence of EPJ during the aerial stage on the robot's orientation.

After the robot step into air, EPJ opens at a certain angle, the leg and rear frame can be regarded as a whole object, with the CoM named $CoM_A$. The body's CoM is denoted as $CoM_B$. Both parts undergo relative rotational motion around $O$ under the tension of the spring. In essence, the robot is simplified into two parts. In order to better analyze the working principle of EPJ, the mechanical structure is simplified, and the results are shown in Figure 2.

\begin{figure}[htbp]
	\centering{\includegraphics[width=0.5\textwidth]{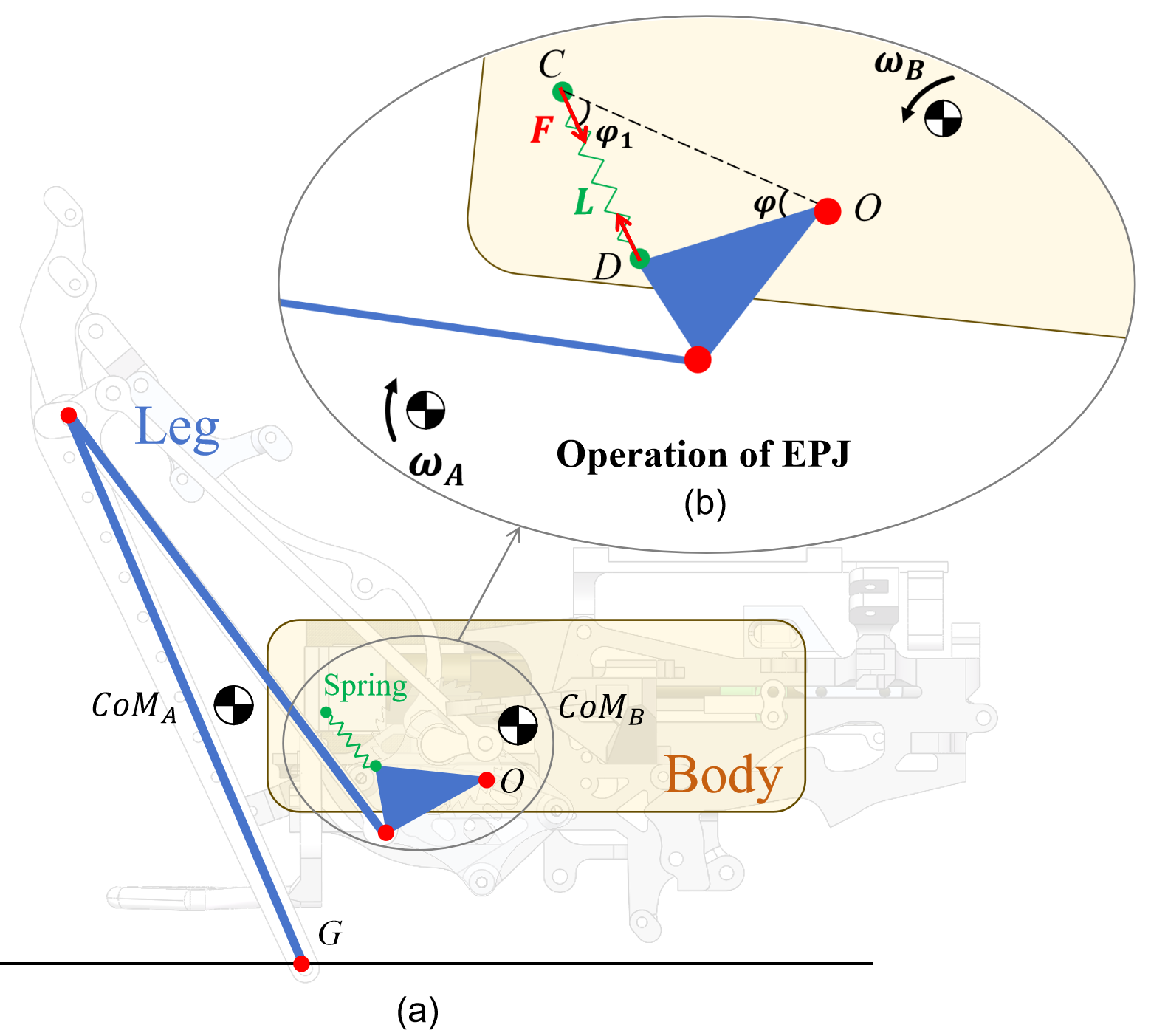}}
	\caption{Simplified Model of Mechanical Structure: (a)EPJ is not triggered and the leg is simplified into a two-link mechanism; (b)EPJ works,  generating a rotation angle between the body and the leg.}
	\label{fig}
\end{figure}


In this paper, $l_{ij}$ represents the distance between points $i$ and $j$, where $\mathrm{i} \& \mathrm{j} \in\{\mathrm{C}, \mathrm{D}, \mathrm{O}\}$. $\omega_p(t)$ denotes the angular velocity of mechanism $p$ at time $t$ (where $p = A,B$). $\varphi(t)$ represents the angle between OC and OD, while $\varphi_1(t)$ represents the angle between OC and spring CD.

According to the cosine theorem of $\varphi(t)$, the length of the spring is
\begin{equation}
	L(t)=\sqrt{l_{\mathrm{OC}}^2+l_{\mathrm{OD}}^2-2 \cdot  l_{\mathrm{OC}} \cdot l_{\mathrm{OD}} \cdot \cos \varphi(t)}
\end{equation}
then, the elastic force $F(t)$ of the spring is
\begin{equation}
	F(t)=k \cdot \Delta L=k \cdot\left[L(t)-L_0\right]
\end{equation}
where $k$ is the spring constant, $L_0$ is the natural length of the spring. The torque exerted by the spring force on mechanisms A and B is
\begin{equation}
	\begin{gathered}
		M(t)=k \cdot l_{\mathrm{OC}} \cdot\left[L(t)-L_0\right] \cdot \sin \varphi_1(t) \\
		\sin \varphi_1(t)=\frac{l_{\mathrm{OD}}}{L(t)} \sin \varphi(t)
	\end{gathered}
\end{equation}
according to the laws of rotation, we can determine that:
\begin{equation}
	\left\{\begin{array}{l}
		M(t)=J_A \frac{d \omega_A(t)}{d t} \Rightarrow \omega_A(t)=\frac{1}{J_A} \int_{t_0}^t M(t) d t+\omega_A\left(t_0\right) \\
		M(t)=J_B \frac{d \omega_B(t)}{d t} \Rightarrow \omega_B(t)=\frac{1}{J_B} \int_{t_0}^t M(t) d t+\omega_B\left(t_0\right)
	\end{array}\right.
\end{equation}
and by the relationship of rotation angles, we can obtain:
\begin{equation}
	\left\{\begin{array}{c}
		\Delta \theta_A=\theta_A(t)-\theta_A\left(t_0\right)=\int_{t_0}^t \omega_A(t) \mathrm{d}t \\
		\Delta \theta_B=\theta_B(t)-\theta_B\left(t_0\right)=\int_{t_0}^t \omega_B(t) \mathrm{d}t \\
		\varphi(t)=\varphi\left(t_0\right)-\Delta \theta_A-\Delta \theta_B
	\end{array}\right.
\end{equation}
if $t=t_{end}$ and EPJ closes, meaning $\varphi(t_{end})=\varphi_0$, then there is boundary condition:
\begin{equation}
	\varphi\left(t_0\right)-\varphi_0=\int_{t_0}^{t_{\text {end }}} \omega_A(t) \mathrm{d}t+\int_{t_0}^{t_{\text {end }}} \omega_B(t) \mathrm{d}t
\end{equation}

Since the kinematic analysis of the jumping phase has already determined the orientations of the leg and body at each moment, therefore at $t_0$, [$\varphi\left(t_0\right), \theta_A\left(t_0\right), \theta_B\left(t_0\right), \omega_A\left(t_0\right), \omega_B\left(t_0\right)$]are all known quantities.

By analyzing the angular velocities of the robot and the rotation angles of each mechanism, we can determine the time $t_{end}$ required for EPJ to close from open. Subsequently, we can solve for the angular velocity of the robot just before EPJ invalidating. At the invalidation moment of EPJ, there will be a collision between the switch and the body, after which the switch will be locked. During this process, angular momentum is conserved. Based on conservation of angular momentum, we can calculate the final angular velocity of the robot, denoted as $\omega_{end}$.




\section{SIMULATION ANALYSIS}

From the previous text, it can be inferred that the switch can control the triggering and invalidating of EPJ. In simulation, we can prevent EPJ from rotating by adding constraints, effectively treating it as if EPJ is not present in the robotic system. In order to better demonstrate the role of EPJ, this chapter conducts simulation verification and comparison of two scenarios: one with EPJ worked and another without EPJ worked. The comparison mainly focuses on the following two aspects: 1) the orientation of the robot before takeoff; 2) the maximum jumping height, distance, and angular velocity of the robot. Additionally, in order to optimize EPJ, an analysis is conducted on the impact of EPJ's position and the spring stiffness coefficient on the final angular velocity, jumping height, and distance of the robot.

The simulation is conducted in Matlab, and the robot model is imported into Simulink through Solidworks for robot simulation, with configuration of the simulation environment. To better simulate the operation of the switch, a weld joint is introduced in Simulink to replace the function of the switch. When the robot starts to jump, the leg gradually opens, and when it reach a certain angle, the weld joint disconnects, opening EPJ. As the robot jumps, the leg gradually rotates to approach the body under the action of the spring. When the rotation angle of EPJ $\varphi(t)=\varphi_0$, the weld joint is re-welded, and the switch closes. For the simulation of robot jumping without EPJ, it is only necessary to keep the weld joint in a welded state, which achieves control of the switch in the simulation.

The parameters of the robot in the simulation are the same as the prototype of the robot. The total mass of the robot is 152.26g, with the leg weighing 33.46g and the body weighing 118.8g. The mass and moments of inertia of each part of the robot are calculated by Solidworks to make the simulation environment more closely resemble the hardware environment.

\subsection{Comparative analysis with or without EPJ}

To demonstrate the improvement of EPJ on the robot's flipping problem, a comparative analysis of the robot's jumping orientation with and without EPJ is required. During the take-off stage, the spring of EPJ stores some energy as elastic potential energy, affecting the orientation of the robot. To visually represent the orientation of the robot, the robot is simplified into a multi-link model, as shown in Figure 3. Figure 3(a) represents the robot model with EPJ, while Figure 3(d) represents the robot model without EPJ.

\begin{figure*}[htbp]
	\centering{\includegraphics[width=1\textwidth]{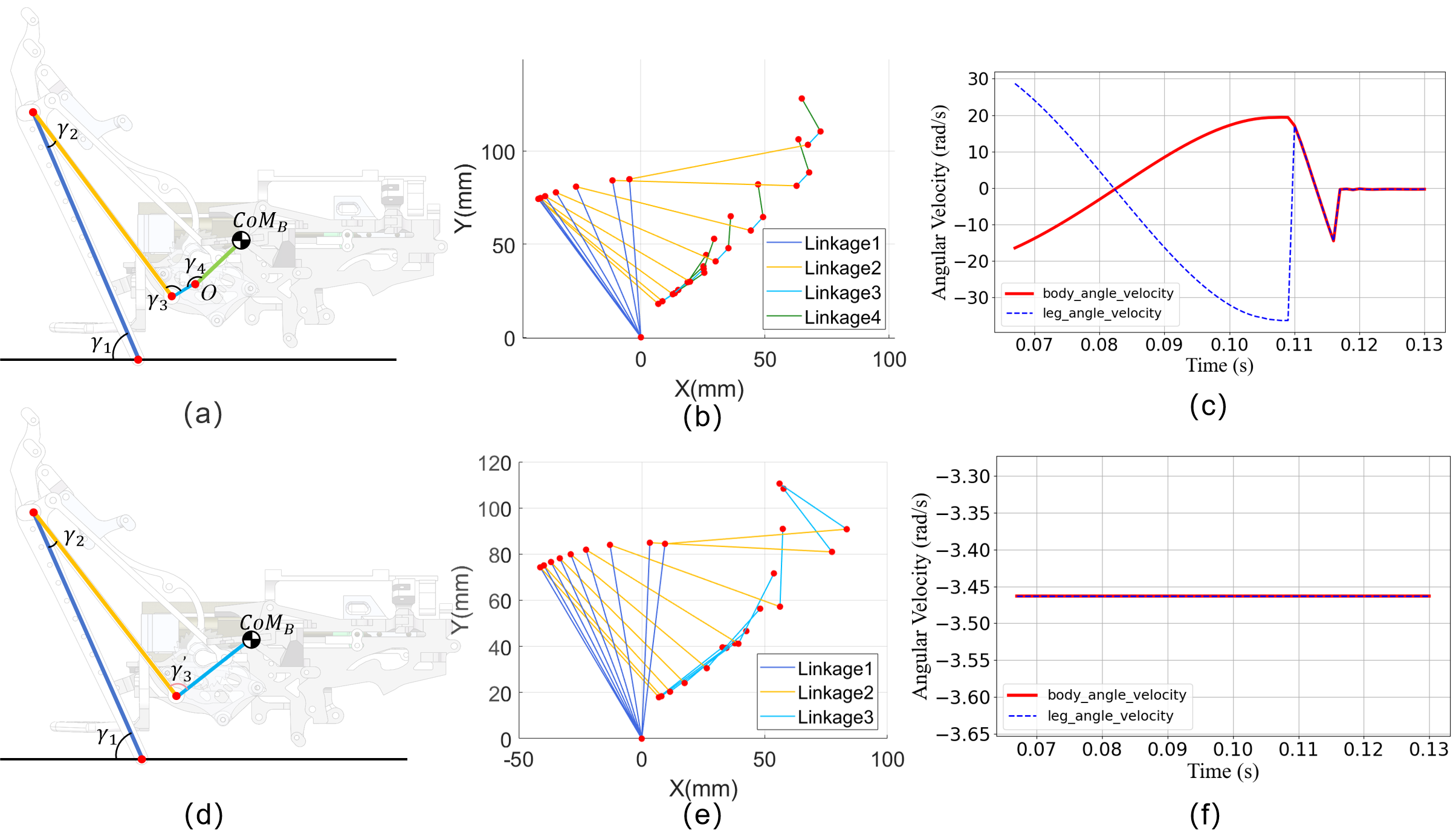}}
	\caption{The jumping orientation of a robot: (a)Robot with EPJ, the robot is simplified to a four-bar linkage; (b)The motion trajectories of each joint with EPJ. At the moment of take-off, the body's angular velocity is -0.02 rad/s, and the leg's angular velocity is 29.99 rad/s; (c)Changes in the robot's angular velocity during the aerial phase; (d)Robot without EPJ. The mechanism is simplified to a three-bar linkage; (e)The motion trajectory without EPJ; (f)After taking-off, the angular velocity is -3.46 rad/s.}
	\label{fig}
\end{figure*}

For the robot with EPJ, its body and leg rotate under the action of the spring after the robot leaving the ground. This process will change the angular velocity of the robot's body and leg, as shown in Figure 3(c). For the robot without EPJ, the leg of the robot is fully extended, and the robot can be considered as a rigid body undergoing projectile motion with constant angular velocity -3.46rad/s, as shown in Figure 3(f).

Simulations are performed for cases with and without EPJ, the joint motion paths and link orientation parameters in the take-off stage are obtained, as shown in Figure 3(b) and Figure 3(e). The simulation results show that when EPJ is working, the robot's final angular velocity is -0.37 rad/s, with a jump height of 49.29 cm and distance of 1.50 m; without EPJ the final angular velocity is -3.46 rad/s, with a jump height of 49.68 cm and distance of 1.46 m. These results indicate that the EPJ can greatly suppress the robot's flipping. 

\subsection{Optimization analysis of the parameter of EPJ}\label{AA}

Different parameters of EPJ have different effects, which will have a certain impact on the angular velocity of the robot. Some important parameters of EPJ mainly include: the position of the revolute joint and the stiffness coefficient of the spring. Among them, the position of the revolute joint directly affects the CoM position, thereby affecting the effect of EPJ. Therefore, we need to analyze how different revolute joint positions will affect the robot. In order to facilitate the description of the position of the revolute joint, the rear end of the robot body part is the zero point of the $x$-axis, and the lowest end of the robot body part is the zero point of the $y$-axis. During the simulation, the positions of the revolute joint in the $x$ and $y$ directions were changed respectively. The simulation results are shown in Figure 4.

\begin{figure}[htbp]
	\centering{\includegraphics[width=0.45\textwidth]{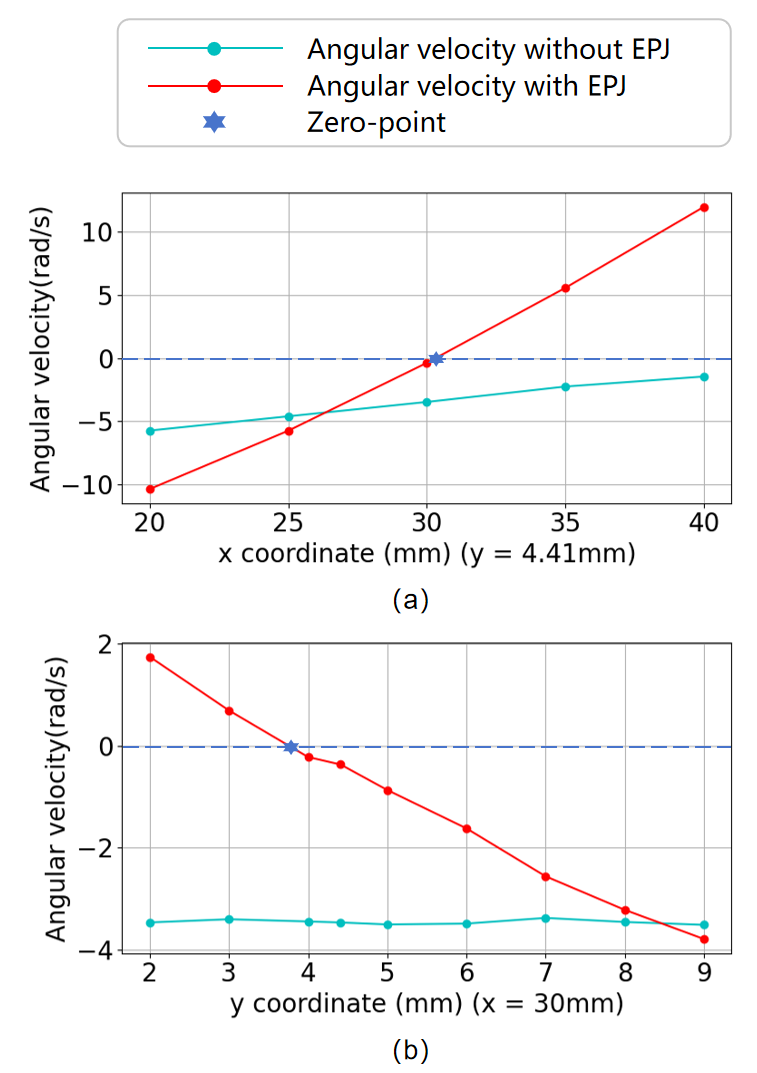}}
	\caption{The relation between the position of the revolute joint and the angular velocity.(a)The angular velocity changes with x coordinate.(b)The angular velocity changes with y coordinate.}
	\label{fig}
\end{figure}

According to Figure 4, the position of the revolute joint has little impact on the angular velocity of the robot without EPJ, and the robot is always in a forward flipping state. When EPJ works, its effect changes with the position of the revolute joint. In Figure 4(a), when the position of the revolute joint is backward, EPJ will increase its forward angular velocity. As the position of the revolute joint gradually approaches the front end, EPJ will cause the robot to flip from front to back. When the x-coordinate is 30.35mm, the robot's angular velocity is 0 rad/s. Regarding the height of the revolute joint as shown in Figure 4(b), EPJ will cause the robot to flip from backward to forward with while height increasing. The angular velocity reaches 0 rad/s when the y-coordinate is 3.7 mm. 

In addition to the position of the revolute joint, the stiffness coefficient of the spring also has a great influence on the effect of EPJ. We simulate spring with different stiffness coefficients and analyze the final angular velocity, jump height and distance of the robot, the results are shown in Figure 5.

\begin{figure}[htbp]
	\centering{\includegraphics[width=0.5\textwidth]{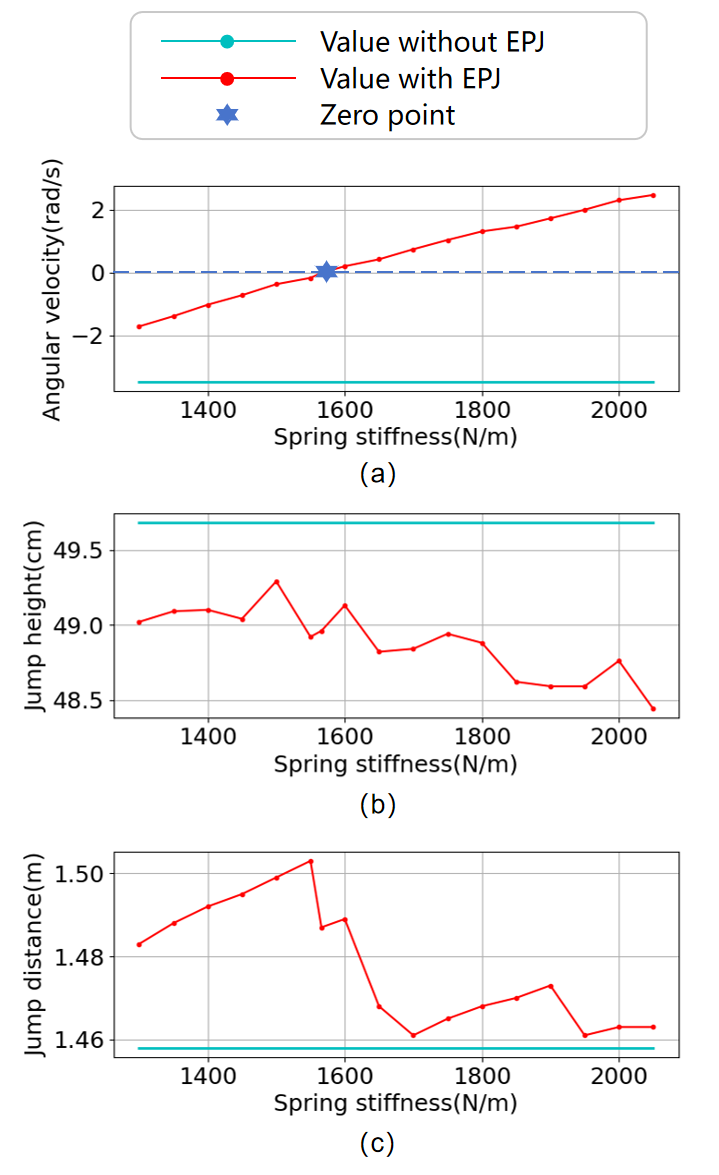}}
	\caption{The results of EPJ parameter optimization.(a)The final angular velocity of the robot varies with the stiffness coefficient of the spring.(b)The jumping height of the robot varies with the stiffness coefficient of the spring.(c)The jumping distance of the robot varies with the stiffness coefficient of the spring.}
	\label{fig}
\end{figure}

The robot without EPJ flips forward at an angular velocity of 3.46 rad/s. When EPJ is working, the tendency of the robot to turn forward is suppressed. With an increase in the stiffness coefficient of the spring, the angular velocity of the robot's forward flipping gradually decreases, as shown in Figure 5(a). When the stiffness coefficient of the spring is 1566 N/m, the robot's angular velocity is 0 rad/s. When the stiffness coefficient of the spring is greater than 1566 N/m, the robot gradually begins to flip backward.

As for the robot's jumping height, the value remains around 49 cm regardless of the variation in the stiffness coefficient of the spring in Figure 5(b). However, the jumping height with EPJ is consistently lower than the height without EPJ. Regarding the robot's jumping distance in Figure 5(c), the greater the stiffness coefficient of the spring, the farther the jumping distance. However, since the jumping distance is measured by the position of the robot's CoM, if the robot lands with poor orientation (such as leaning backward), the position of the CoM will be further back, resulting in a shorter jumping distance. 

In summary, when the position of EPJ is close to the upper and rear part, the forward flipping will be intensified. When it is close to the lower and front part, the robot will gradually transition from forward flipping to backward flipping. In this process, there's a position where the angular velocity is 0 rad/s. For the stiffness coefficient of the spring, with the value gradually increasing, the robot transition from forward flipping to backward flipping. Additionally, regardless of how the stiffness coefficient varies, EPJ always leads to a slight sacrifice in jumping height while increasing the jumping distance.

\section{CONCLUSIONS}

This paper designs a mechanism EPJ composed of a revolute joint, a switch and a spring, and conducts dynamic modeling and analysis of the jumping robot with EPJ. In order to verify the effect of EPJ, virtual prototype modeling and jumping simulation experiments are implemented in MATLAB , comparing the jumping parameters between with and without EPJ. The results show that the angular velocity of the robot without EPJ is -3.46rad/s, and the one with EPJ is -0.37rad/s. It indicates that EPJ can effectively suppress the flipping problem of jumping robots. 

By optimizing the parameters of EPJ, it is found that the effect of EPJ on the robot is different depending on the position of the revolute joint. When the position of EPJ is close to the back end, it will intensify the forward flipping of the robot; on the other hand, when it is close to the front, it will gradually change the robot from forward flipping to backward flipping. During this process there will be a point where the angular velocity is 0 rad/s, representing the correct position that EPJ effectively eliminates the flip during the jump. In addition, the optimal spring stiffness coefficient is 1566N/m, at which value the robot can achieve a complete non-flipping jump. 


\addtolength{\textheight}{-12cm}  





\bibliographystyle{IEEEtran}
%

\end{document}